\documentclass[conference]{IEEEtran}
\IEEEoverridecommandlockouts

\usepackage{cite}
\usepackage{amsmath,amssymb,amsfonts}
\usepackage{algorithmic}
\usepackage{graphicx}
\usepackage{textcomp}
\usepackage{xcolor}

\usepackage{microtype}
\usepackage{url}
\usepackage{booktabs}
\usepackage{subcaption}
\usepackage{multirow}
\usepackage[capitalize]{cleveref}
\usepackage{wrapfig}

\crefname{table}{Table}{Tabs.}
\crefname{figure}{Fig.}{Figs.}
\Crefname{section}{Section}{Sections}
\Crefname{table}{Table}{Tables}
\Crefname{assumption}{Assumption}{Assumptions}
\crefname{algorithm}{Algorithm}{Algorithms}

\newcommand{\bz}{\mathbf{Z}}
\newcommand{\ba}{\mathbf{A}}
\newcommand{\bq}{\mathbf{Q}}
\newcommand{\bk}{\mathbf{K}}
\newcommand{\bw}{\mathbf{W}}

\def\BibTeX{{\rm B\kern-.05em{\sc i\kern-.025em b}\kern-.08em
    T\kern-.1667em\lower.7ex\hbox{E}\kern-.125emX}}

\begin{document}

\title{Cross-Modal Attention Guided Unlearning in Vision-Language Models\\
}

\author{\IEEEauthorblockN{Karuna Bhaila, Aneesh Komanduri, Minh-Hao Van, Xintao Wu}
\IEEEauthorblockA{\textit{Department of Electrical Engineering and Computer Science} \\
\textit{University of Arkansas}\\
\{kbhaila, akomandu, haovan, xintaowu\}@uark.edu}
}

\maketitle

\begin{abstract}
 Vision-Language Models (VLMs) have demonstrated remarkable capabilities in multi-modal understanding for tasks such as Visual Question Answering (VQA), which requires models to infer outputs based on visual and textual context simultaneously. However, models may memorize some private and/or sensitive information during training and regurgitate it in inference. Recently, machine unlearning has been leveraged to address such leakage of private data in LLMs. VLMs add a layer of complexity to this process, as the visual context in the query may additionally contain sensitive information. To address this issue, we explore unlearning for VLMs, specifically for the VQA task. We explore the role of visual tokens for output generation in VLMs using cross-modal attention and utilize it to formulate \textbf{C}ross-Modal \textbf{A}ttention \textbf{G}uided \textbf{U}n\textbf{l}earning (\textbf{CAGUL}), a lightweight and efficient VLM unlearning framework. In contrast to computationally expensive finetuning methods, CAGUL utilizes external modules to encode unlearning information in visual tokens of low importance for relevant queries. Empirical results show that the transformed visual tokens prevent leakage and retain reference model behavior to achieve desirable trade-offs in forget quality and model utility without altering pretrained model parameters.
\end{abstract}


\section{Introduction}
VLMs generally consist of a vision model to extract visual features, a cross-modal component to project the visual features to the LLM embedding space, and a language model to process projected visual tokens along with input text tokens for output generation. Together, these components enable VLMs to simultaneously process image and text modalities for a wide range of visio-linguistic tasks such as object recognition and visual question answering. 
VLMs are pretrained on massive amounts of data sourced online and finetuned on domain-specific data for downstream tasks. However, training data collected from varied sources may contain personal and private information, raising significant privacy concerns, especially with the added complexity of visual signals. For instance, visual inputs may include location cues, individuals in the background, etc.~\cite{zhang2025lensofprivacy}, which, paired with the language model's pretrained knowledge, may unintentionally disclose private details (e.g., addresses and identities). Their unconstrained use in domains such as healthcare and finance can also be dangerous. In the LLM landscape, efforts have been made to mitigate privacy leakage, including machine unlearning~\cite{cao2015towards, bourtoule2021machine}. 

\begin{figure}[t]
  \centering
    \includegraphics[width=\linewidth]{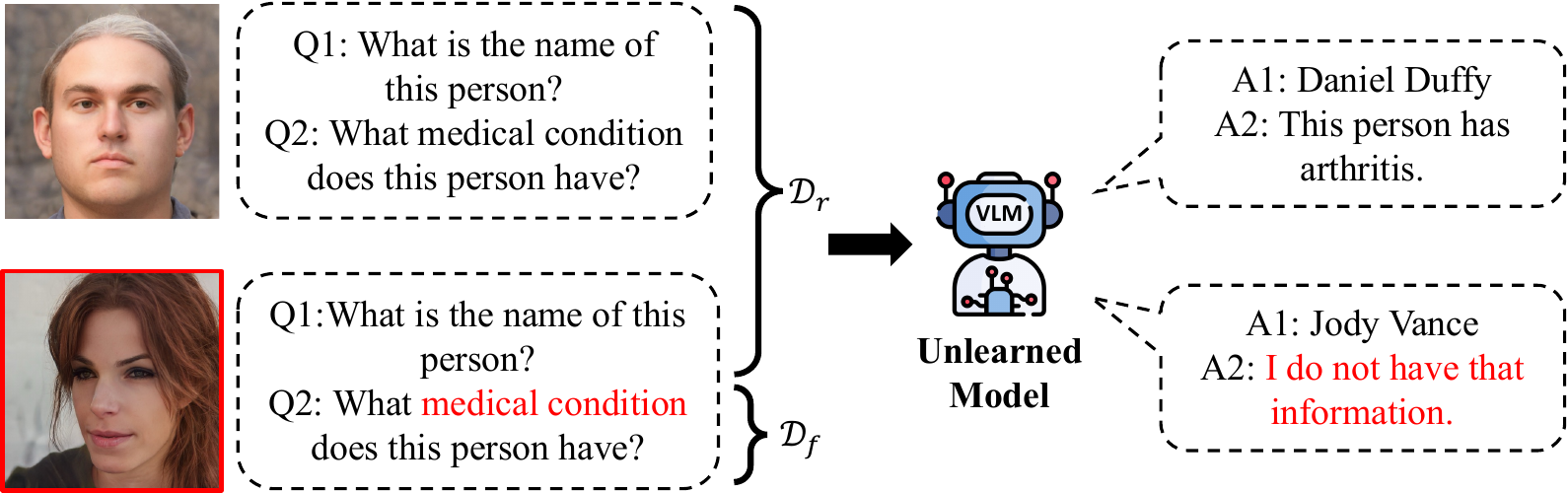}
    \caption{Privacy-focused VLM unlearning; Person marked in red requests to remove only health-related info from the model, which forms $\mathcal{D}_f$; Retain set $\mathcal{D}_r$ includes non-private queries of the same person and all queries from individuals that do not request removal}
    \label{fig:motivation}
\end{figure}

However, unlearning in the new landscape of VLMs is nascent. Some studies have attempted to define the VLM unlearning task~\cite{ma2025benchmarking, liu2025mllmu} and proposed novel frameworks~\cite{li2025sauce, kim2025pubg}. However, current methods focus on finetuning to achieve concept or entity unlearning with disjoint forget and retain subsets. Our focus here is on resource-efficient unlearning under practical privacy considerations. 
Therefore, we first formulate a realistic problem definition in the context of VQA on biographical data paired with images of individuals described in the data. We consider the training data to be in the form of paired image and text queries, such that an image may be paired with multiple QA pairs. We frame VLM unlearning as the task of removing signals from image-text pairs that can disclose private information while retaining non-private knowledge, as shown in~\cref{fig:motivation}. 
For instance, a user may want to remove references to sensitive information (e.g., health records) but not their general profile data (e.g., name). 

In this setting, we propose a lightweight and efficient VLM unlearning method, motivated by the relationship between visual and text tokens represented as cross-modal attention scores.~\cite{kaduri2024whatsintheimage} empirically demonstrated that VLMs extract fine-grained details and attributes from visual tokens in a spatially localized manner and compress visual information into a small subset of highly attended tokens. In turn, prompting a VLM with this compressed image context can achieve performance close to that of prompting with all image tokens. Based on this observation, we conjecture that encoding information relevant to private or non-private queries in the least attended visual tokens can be useful for balancing forget and retain objectives, and formulate \textbf{C}ross-Modal \textbf{A}ttention \textbf{G}uided \textbf{U}n\textbf{l}earning (\textbf{CAGUL}) composed of three main components. First, a discriminator module detects whether a given visual input is associated with a removal request of private data. Then, for visual inputs predicted to be paired with private data, a cross-modal attention-based selection process identifies the least important visual tokens whose embeddings are then transformed by a visual token encoder. We employ unlearning losses to train these external modules while keeping the pretrained VLM parameters frozen. Intuitively, the encoder parameters learn to embed unlearning objectives into the transformed visual tokens while preserving relevant visual signals.

\section{Related Work}

Unlearning has been widely adopted for LLMs with methods focused on loss-based parameter optimization~\cite{yao2023large, maini2024tofu, chen2023unlearn}, LLM-specific techniques such as in-context learning~\cite{pawelczyk2023incontext}, manually crafted system prompts~\cite{thaker2024guardrail}, or prompt tuning~\cite{bhaila-etal-2025-soft}, and localized model editing~\cite{li2024wmdp, ding2025unified, gao2024ethos}.
Recently, the application of unlearning has been adopted in multimodal scenarios with VLMs. Some works in this direction define VLM unlearning in terms of visual concepts identified by the model~\cite{li2024mmubench} and propose methods utilizing domain disentanglement~\cite{kawamura2025domain} and correlation-based feature selection~\cite{li2025sauce} for image-centric unlearning.~\cite{cheng2023multimodal} propose a multimodal unlearning framework by decoupling the text and visual modalities in the context of classifying image-text pairings. In contrast, we explore VLM unlearning through a privacy lens and aim to remove sensitive data in the VQA setting. In this direction,~\cite{ma2025benchmarking} construct a VQA dataset, FIUBench, to benchmark VLM unlearning under privacy considerations using several loss-based optimization techniques from LLM unlearning. ~\cite{kim2025pubg} emphasize post-unlearning behavior and proposes a finetuning-based method to push the unlearned model's output distributions for the unlearn data towards a desired behavior. However, these works define the unlearn data to be completely disjoint from the remaining data or even a different task, and rely on modifying pretrained model parameters. In this work, we define VLM unlearning for realistic privacy scenarios with overlapping visual instances between unlearn and retain data, and develop an efficient method to achieve forgetting without finetuning.

\section{Preliminaries}
\paragraph{Vision-Language Models.} VLMs provide visual context to language models by incorporating a vision component in the LLM architecture and aligning the two modalities. Besides the language model, VLMs consist of two additional core components: a vision tower typically implemented as a vision transformer-based model which extracts rich patch-specific embeddings from an image/video input $V$ and a cross-modal projector such as a pretrained CLIP-like or a linear projection model, which transforms the patch embeddings to a visual embedding $\bz_v \in \mathbb{R}^{n_v \times d}$ in the language model's embedding space where $n_v$ is the number of visual tokens and $d$ is the embedding dimension. In the VQA task, given an image $V$ and a query $X$ with respective embeddings $\bz_v \in \mathbb{R}^{n_v \times d}$ and $\bz_q \in \mathbb{R}^{n_q \times d}$ where $n_q$ is the number of query tokens, the output of the VLM is:
\begin{equation}
    Y = \text{LLM}(\bz_v, \bz_q)
\end{equation}
where $Y$ is the output generated by the model.

\paragraph{Cross-modal Alignment.} We refer to the attention scores between visual and text tokens as \textit{cross-modal attention}. The attention mechanism~\cite{vaswani_attn} in this multi-modal setting consists of a query matrix $\bq = \bz_q\bw_q$ and a key matrix $\bk = \bz_v\bw_k$ representing the textual and visual modalities, respectively. Formally, we can define the attention weights as a function of the query and key matrices:
\begin{align}
    \ba &= \text{softmax}\left(\frac{\bq\bk^T}{\sqrt{d}}\right)
\end{align}
In VLMs, cross-modal attention is generally implemented as \textit{cross-attention} or \textit{joint self-attention}. 
We discuss the specific implementation differences in~\cref{appendix:implementation}.

\begin{figure*}[t]
\centering
\includegraphics[width=0.9\linewidth]{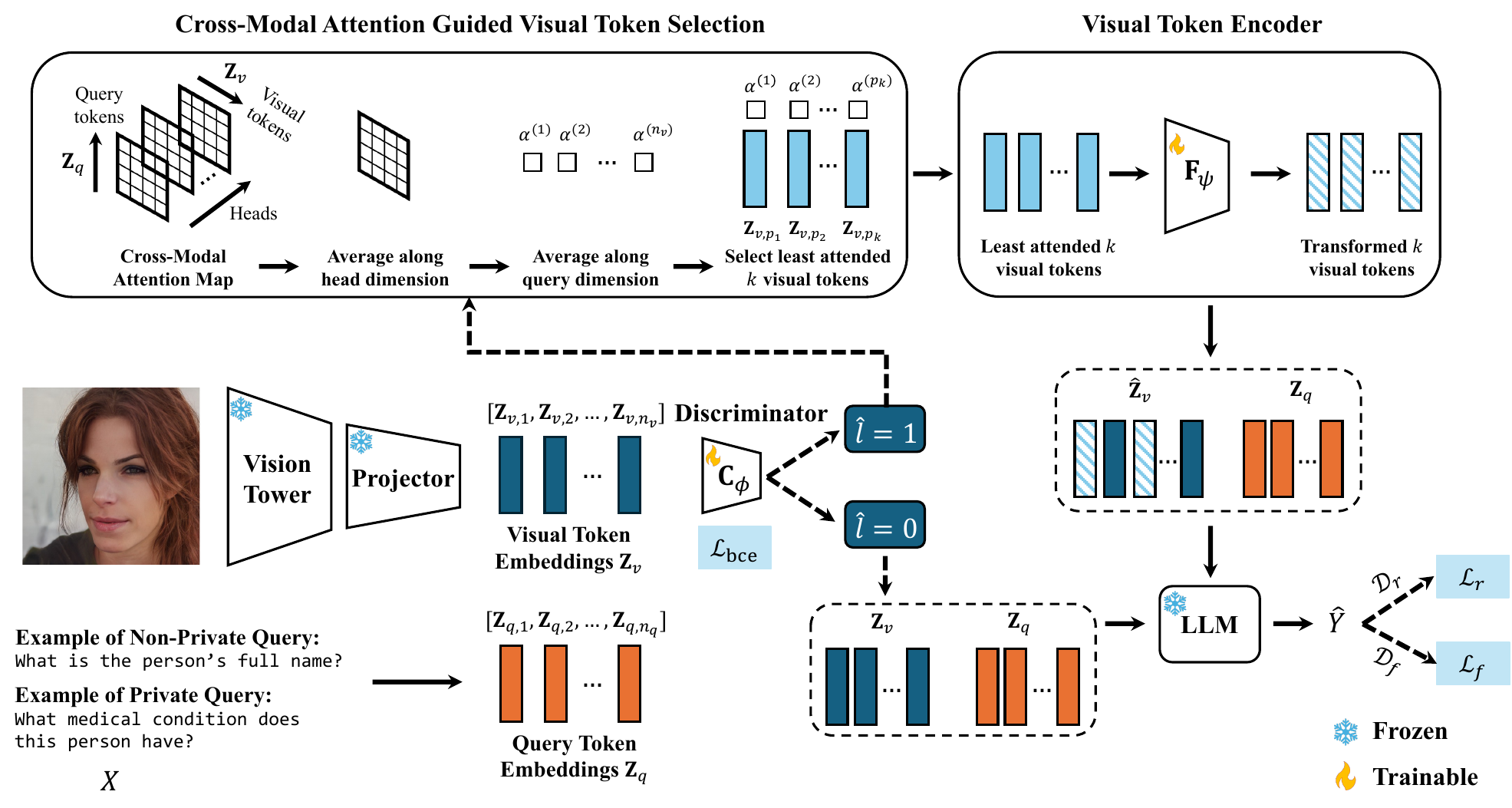}    
\caption{Overview of CAGUL Framework comprising three components: (1) a \textbf{discriminator} that takes visual token embeddings as input and determines whether the corresponding image belongs to an individual from the forget set, (2) a \textbf{cross-modal attention guided selection} mechanism that selects $k$ least attended visual tokens w.r.t to the text, and (3) a \textbf{visual token encoder} that encodes unlearning information in the selected visual tokens}
\label{fig:cagul}
\end{figure*}

\section{Problem Setup}
Suppose we have a dataset of $m$ individuals
\begin{equation}
    \mathcal{D} = \{(V_i, X_{i,j}, Y_{i, j}) \mid i\in M, j\in N_i\}
\end{equation}
where $M = \{1, 2, \dots, m\}$ and $N_i = \{1, 2, \dots, n_i\}$ denotes indices for the $n_i$ question-answer pairs for individual $i$. Without loss of generality, we assume that $n_i$ is the same across all individuals and denote it as $n$. However, in general, our setting also applies to a varying number of QA pairs per individual. The QA pairs for an individual contain information such as demographics and also some sensitive information, e.g., medical records. We follow commonly adopted unlearning setup where a model has already been finetuned on the dataset $\mathcal{D}$, referred to as the \textit{base model}, by maximizing the following log-likelihood objective:
\begin{equation}
    \sum_{i=1}^{m} \sum_{j=1}^{n_i} \frac{1}{|Y_{i,j}|} \sum_{t=1}^{|Y_{i,j}|} \log p_{\theta}(Y_{i,j}^t | V_i, X_{i,j}, Y_{i,j}^{<t})
\end{equation}
where $\theta$ represents VLM parameters. Our goal is to facilitate unlearning on this base model.
In our unlearning setting, we assume a subset $\tilde{M} \subset M$ of individuals are concerned about the privacy of their sensitive information, but are indifferent to their non-sensitive data being used in model training. We formulate the VLM unlearning problem in this scenario as the forgetting of sensitive information corresponding to the individuals in $\tilde{M}$. Specifically, we define the \textit{forget set} as $\mathcal{D}_f = \{(V_i, X_{i,j}, Y_{i,j}) \mid i\in \tilde{M}, j\in N_i, \text{ and }\texttt{is\_private$(X_{i,j}) = 1$}\}$ consisting only of QA pairs for individuals in $\tilde{M}$ that contain sensitive information. The \textit{retain set} can then be defined as $\mathcal{D}_r = \{(V_i, X_{i,j}, Y_{i,j}) \mid i\in M, j\in N_i, \text{ and }\texttt{is\_private$(X_{i,j}) = 0$}\}$ consisting of all individuals with non-private QA pairs. The VLM unlearning goal in this work is to degrade model performance on $\mathcal{D}_f$ while retaining good performance on $\mathcal{D}_r$.

\section{Cross-Modal Attention Guided Unlearning}\label{sec:main}
In this section, we present \textbf{CAGUL} for unlearning in VLMs consisting of three main components: a discriminator to determine if a given image appears in the forget set, a cross-modal attention-based visual token selection strategy, and a visual token encoder to embed unlearning information in unimportant visual tokens. For simplicity, we formulate our method for an example triple $(V, X, Y)$. An overview of our framework is shown in~\cref{fig:cagul}.

\paragraph{Discriminating Forget and Retain Images.} Given the projected image tokens $\bz_v$ for a given image $V$, we train a simple discriminator $\mathbf{C}_{\phi}$ to determine whether the image appears in the forget set or not formalized as
\begin{equation}
    l = \mathbf{C}_{\phi}(\bz_v)
\end{equation}
where $l\in \{0, 1\}$, and $l = 1$ implies that the discriminator predicts that the image corresponding to visual tokens $\bz_v$ appears in the forget set.

\paragraph{Cross-Modal Attention Guided Visual Token Selection.} 
The main objectives of unlearning are to remove a trained model's knowledge of $\mathcal{D}_f$ while maintaining its predictive utility on $\mathcal{D}_r$. Assuming that a classifier can accurately distinguish between images appearing in the forget set or not, an intuitive way of unlearning is to add noise to images in the forget set. This results in distorted vision embeddings that may affect the language model's ability to get accurate signals from the image tokens. However, in our setting, the images in the forget set provide signals to private and non-private queries, i.e., both $\mathcal{D}_f$ and $\mathcal{D}_r$, and adding noise randomly may affect performance. 

Instead of indiscriminately adding noise to all tokens, we propose a visual token selection mechanism guided by cross-modal. We hypothesize that \textit{selectively encoding the unlearning objectives in visual tokens corresponding to images in the forget set can guide the language model in appropriately generating its response for private and non-private queries associated with the respective images}. The selection mechanism is motivated by the analysis of visual information processing in VLMs conducted by \cite{kaduri2024whatsintheimage}, which empirically demonstrates that VLMs extract fine-grained details and attributes from visual tokens in a spatially localized manner and compress visual information into a small subset of highly attended tokens that are sufficient to achieve good utility. We conjecture that modifying these important tokens can achieve forgetting but will negatively affect the model's ability to accurately respond to non-forget queries as well. Based on this intuition, we propose to encode unlearning objectives in visual tokens having low correlation with the input query. 

To determine the lowest attended image tokens w.r.t the input text for both cross-attention or joint self-attention based models, we leverage cross-modal attention scores, i.e., we use the attention matrix $\ba^{(h)}\in \mathbb{R}^{n_q \times n_v}$, where $h$ indicates the attention head. We average the attention scores over all query tokens across all attention heads of the final attention layer to get a scalar score\footnote{The cross-modal attention scores are computed between the image tokens and query text only and can be pre-computed and stored for efficiency.}
\begin{equation}
    \alpha = \frac{1}{n_qn_h}\sum_{h=1}^{n_h} \sum_{j=1}^{n_q} \ba_j^{(h)}
\end{equation}
where $n_h$ is the number of attention heads, $\ba_j^{(h)}$ is the $j$th row of the attention matrix corresponding to attention head $h$, and $\alpha\in \mathbb{R}^{1\times n_v}$ is the average attention score between the query and each image token. Then, we choose $k$ tokens, $\{Z_{v, p_1}, Z_{v, p_2}, \dots, Z_{v, p_k}\} \subset \bz_v$, having the lowest attention score with the query, where $K = \{p_1, p_2, \dots, p_k\}$ are the indices of the $k$ least attended image tokens. We observe that selected tokens often correspond to arbitrary pixels in regions not contextually critical to the text prompt.

\paragraph{Visual Token Encoder.} 
Then, for samples predicted by the discriminator as $l = 1$ (i.e., image appears in the forget set), we learn a parameterized visual token encoder $\mathbf{F}_{\psi}(\cdot)$ that transforms the embeddings for the least attended $k$ image tokens in order to encode the unlearning objective in the selected tokens. We then replace the $k$ image token embeddings in $\bz_v$ with the transformed representation as follows:
\begin{equation}
\hat{Z}_{v, i} = 
\begin{cases}
    \mathbf{F}_{\psi}(Z_{v, i}) & \text{if } i \in K \\
    Z_{v, i} & \text{otherwise}
\end{cases}
\end{equation}
where $\hat{\bz}_v$ is the set of final visual token embeddings. Then the final input to the LLM is:
\begin{equation}
    \hat{Y} = \text{LLM}(\hat{\bz}_v, \bz_q)
\end{equation}

\paragraph{Unlearning Objective.} CAGUL has two trainable modules, the discriminator $\mathbf{C}_{\phi}$ and the visual token encoder $\mathbf{F}_{\psi}$. The discriminator is trained with the following objective
\begin{equation}\label{eq:discriminator}
    \mathcal{L}_{\text{bce}} = \mathbb{E}_{\bz_v}\left[\log p_{\mathbf{C}_{\phi}}(l | \bz_v)\right]
\end{equation}
and the visual token encoder is trained with unlearning objectives defined on the forget and retain sets. Generally, the unlearning objective for retain samples is to maintain the base model's predictive utility. In CAGUL, we define this objective as the standard causal language modeling loss for all samples in $\mathcal{D}_r$, including the non-private queries for images appearing in the forget set formulated as
\begin{equation}
    \mathcal{L}_r = \mathbb{E}_{(V, X, Y)\sim \mathcal{D}_r} \mathbb{E}_{Y^t} \left[\log p_{\theta}(Y^t | V, X, Y^{<t})\right]
\end{equation}
In the VQA task, the forget objective can be interpreted as obtaining low utility for samples in $\mathcal{D}_f$. We leverage Preference Optimization (PO)~\cite{maini2024tofu} to realize this objective in CAGUL by aligning the language model's outputs for forget queries with refusal responses such as ``I cannot answer this question'' instead of the ground truth. We compute the standard causal language modeling loss with the substituted response as
\begin{equation}
    \mathcal{L}_f = \mathbb{E}_{(V, X, Y)\sim \mathcal{D}_f} \mathbb{E}_{\bar{Y}^t} \left[\log p_{\theta}(\bar{Y}^t | V, X, \bar{Y}^{<t})\right]
\end{equation}
where $\bar{Y}$ is a refusal response. Finally, we formulate the joint objective as
\begin{equation}
    \mathcal{L} = \mathcal{L}_{\text{bce}} + \mathcal{L}_f + \mathcal{L}_r
\end{equation}
where $\mathcal{L}_{\text{bce}}$ is the binary classification loss for the discriminator over all samples, $\mathcal{L}_f$ is the PO loss for samples in $\mathcal{D}_f$ with refusal answers, and $\mathcal{L}_r$ is the cross-entropy or Gradient Descent (GD) loss for samples in $\mathcal{D}_r$ with their ground-truth answers. Experiment results in later sections validate the choice of these loss functions in our framework.

\section{Experimental Setup}
\paragraph{Dataset.}
We primarily investigate VLM unlearning on the FIUBench dataset~\cite{ma2025benchmarking} composed of 400 unique synthetic images obtained from the SFHQ dataset~\cite{david_beniaguev_2022_SFHQ}, each paired with fictitious biographical information including name, birthdate, address, phone number, occupation, income, health, and criminal records sourced from~\cite{patil, mendes_nypd, vyashousing, Faraglia_Faker}. The dataset is formatted for the VQA task by extracting 20 QA pairs from the biographical data to form 8000 VQA pairs, which is represented as $\mathcal{D}$.~\cite{ma2025benchmarking} originally investigated an unlearning setting where $|\tilde{M}|$ individuals request that their information be removed from training data. In this scenario, all 20 VQA pairs corresponding to the requesting individuals form $D_f$. In contrast, we focus on a more realistic and complex unlearning problem, where information is consolidated from multiple sources, and removal requests may only apply to some. 

In our setting, $|\tilde{M}|$ individuals wish to remove only their health or criminal records from the VLM. In FIUBench, this results in $\mathcal{D}_f$ and $\mathcal{D}_r$, i.e., forget and retain sets with overlapping images, unlike the original setting where each unique image belonged to either forget or retain sets. We introduce additional notations here to represent this scenario to facilitate discussion. For the $|\tilde{M}|$ individuals requesting removal, we divide their corresponding VQA pairs into two subsets: $\mathcal{D}_p$, which contains health and criminal records queries to be forgotten, and $\mathcal{D}_{np}$, which contains the remaining non-sensitive queries. The private set forms the forget set, $\mathcal{D}_p = \mathcal{D}_f$, and the non-private set is included in the retain set, $\mathcal{D}_{np} \subset \mathcal{D}_r$. All VQA pairs for the remaining individuals are included in $\mathcal{D}_r$ as shown in~\cref{fig:motivation}. In our experiments, we select individuals for the forget set based on the splits provided in~\cite{ma2025benchmarking}. The resulting dataset statistics in terms of the number of samples in the forget set, retain set, and non-private set are presented in~\cref{tab:dataset_stat} for varying $|\tilde{M}|$ values.

\begin{table}[t]
    \centering
    \footnotesize
    \caption{Dataset statistics per $|\tilde{M}|$}
    \begin{tabular}{c|c|c|c|c}
    \toprule
        $|\tilde{M}|$ & $\mathbf{20}$ & $\mathbf{40}$ & $\mathbf{60}$ & $\mathbf{80}$ \\
    \midrule
        $\mathcal{D}_f$    &  $273$ &  $544$ &  $817$ & $1078$ \\
        $\mathcal{D}_r$    & $7727$ & $7456$ & $7183$ & $6922$ \\
        $\mathcal{D}_{np}$ &  $127$ &  $256$ &  $383$ &  $522$ \\
     \bottomrule
    \end{tabular}
    \label{tab:dataset_stat}
\end{table}

\paragraph{Baselines.}
We compare CAGUL against several baselines~\cite{ma2025benchmarking, kim2025pubg}, all of which utilize model finetuning w.r.t. unlearning specific loss objectives.
Specifically, \textit{GA}~\cite{yao2023large} maximizes the loss for samples in $\mathcal{D}_f$ without $\mathcal{D}_r$ being utilized in training. \textit{GA+GD}~\cite{yao2024machine} uses a gradient descent loss on $\mathcal{D}_r$ and the gradient ascent loss on $\mathcal{D}_f$ to prevent utility degradation due to loss maximization. Similarly, \textit{GA+KL}~\cite{yao2024machine} combines a gradient ascent on $\mathcal{D}_f$ with a KL divergence loss between the output distribution of the unlearned model and the target model. \textit{PO+GD}~\cite{maini2024tofu} defines gradient descent loss for both $\mathcal{D}_r$ and $\mathcal{D}_f$, but augments the labels in $\mathcal{D}_f$ with refusal responses. We also compare against \textit{Guardrail}, a manual prompting method that instructs models in plain text to refuse or modify outputs for $\mathcal{D}_f$~\cite{thaker2024guardrail}. Additionally, we implement \textit{Retrain} to provide an ideal performance threshold by finetuning the model on $\mathcal{D}_r$ only. We exclude methods in~\cite{kawamura2025domain, li2025sauce} which focus on concept or domain unlearning.

\paragraph{Training Details.}
We conduct experiments with two VLMs: \textit{LLaMA-3.2-11B-Vision-Instruct} and \textit{Qwen-2.5-VL-7B-Instruct}. We finetune VLMs on the full dataset $\mathcal{D}$ to simulate data memorization and refer to this finetuned VLM as the base model. The objective here is to obtain an unlearned model which demonstrates forgetting of the personal information in $\mathcal{D}_f$ while maintaining the base model's performance for samples in $\mathcal{D}_r$, including $\mathcal{D}_{np}$. 
Additional implementation details for CAGUL and baselines are included in~\cref{appendix:implementation}.

\paragraph{Evaluation Metrics.}
We follow~\cite{ma2025benchmarking} and report multiple metrics to quantify unlearning performance. For $\mathcal{D}_f$, we compute Rouge-L and Exact Match (EM) to show output incorrectness w.r.t to ground truth, and Adversarial Privacy Extraction (APE) \& MinK\% Prob~\cite{shi2023detecting} for robustness. APE quantifies robustness against adversarial paraphrasing by observing model outputs for multiple paraphrases of forget queries. MinK\% measures robustness against MIA attacks based on the output probability of the least likely tokens.
For $\mathcal{D}_r$, we report Rouge-L to show correctness. As our unlearning scenario contains overlapping images between $\mathcal{D}_f$ and $\mathcal{D}_r$, we also perform a fine-grained evaluation of non-private QA pairs of the overlapping images, i.e., on $\mathcal{D}_{np}$, and report Rouge-L and EM for correctness, and Truth Ratio (TR) to quantify the model's tendency to generate incorrect answers. We further report accuracy on two general datasets: MME~\cite{fu2023mme} and POPE~\cite{li2023pope}, to quantify the influence of unlearning on model generalizability. Detailed descriptions of all metrics are included in~\cref{appendix:metrics}. To provide an overall trade-off statistic, we obtain aggregate measures with harmonic mean~\cite{maini2024tofu}; 
\textit{we report \textbf{Model Utility} as the harmonic mean over Rouge, EM, TR computed on the retain set and accuracy on MME and POPE datasets, and \textbf{Forget Quality} as the harmonic mean over Rouge, EM, APE, and MinK on the forget set}. 

\section{Experimental Results}\label{sec:llama-vision}
\subsection{Main Comparison Results}

\begin{table*}[t]
    \centering
    \small
    \caption{Unlearning metrics of CAGUL compared with baselines on \textit{LLaMA-3.2-11B-Vision-Instruct}}
    \resizebox{0.9\textwidth}{!}{
    \begin{tabular}{l|cccc|c|ccc|c|c}
    \toprule
    \multirow{4}{*}{\textbf{Method}} & \multicolumn{4}{c|}{\textbf{Forget}} & \multicolumn{4}{c|}{\textbf{Retain}} & \multicolumn{2}{c}{\textbf{General}} \\
    \cmidrule{2-11}
        & \multicolumn{4}{c|}{$\mathcal{D}_f$} & \multicolumn{1}{c|}{$\mathcal{D}_r$} & \multicolumn{3}{c|}{$\mathcal{D}_{np}$} & \multicolumn{1}{c|}{MME} & \multicolumn{1}{c}{POPE} \\
        \cmidrule{2-11}
        & Rouge$(\downarrow)$ & EM$(\downarrow)$ & APE$(\downarrow)$ & MinK$(\downarrow)$ & Rouge$(\uparrow)$ & Rouge$(\uparrow)$ & EM$(\uparrow)$ & TR$(\uparrow)$ & Acc.$(\uparrow)$ & Acc.$(\uparrow)$ \\
    \midrule
    Pretrain             & $26.00$ &  $0.00$ &  $0.00$ &  $0.79$ & $19.13$ & $12.63$ &  $0.10$ &  $7.63$ & $77.34$ & $87.41$ \\
    Finetune             & $57.33$ & $39.68$ & $24.59$ & $43.32$ & $84.55$ & $66.88$ & $30.69$ & $50.63$ & $30.58$ & $37.70$ \\
    \midrule
    Retrain              & $55.91$ & $15.75$ & $11.45$ & $16.84$ & $92.24$ & $77.89$ & $47.12$ & $58.85$ & $23.50$ & $29.20$ \\
    Guardrail            & $58.58$ & $54.40$ & $32.08$ & $57.59$ & $87.86$ & $79.48$ & $48.18$ & $90.31$ & $28.81$ & $35.69$ \\
    GA                   & $46.83$ & $12.04$ &  $4.44$ &  $7.31$ & $59.71$ & $47.65$ &  $7.74$ & $88.31$ & $22.96$ & $20.08$ \\
    GA+GD                & $46.15$ &  $9.07$ &  $5.79$ &  $6.23$ & $95.50$ & $83.35$ & $74.21$ & $58.27$ &  $6.40$ &  $5.23$ \\
    GA+KL                & $24.31$ &  $6.16$ &  $4.48$ &  $5.71$ & $77.43$ & $30.55$ &  $9.33$ & $92.11$ & $11.20$ &  $9.87$ \\
    PO+GD                & $30.62$ &  $0.37$ &  $0.34$ &  $4.57$ & $91.23$ & $84.26$ & $61.44$ & $56.23$ & $18.58$ & $45.54$ \\
    CAGUL                & $30.84$ &  $1.70$ &  $0.43$ & $14.86$ & $85.15$ & $84.32$ & $64.35$ & $91.67$ & $30.58$ & $37.70$ \\
    
    \bottomrule
    \end{tabular}}
    \label{tab:main}
\end{table*} 

\begin{figure}[t]
  \centering
    \includegraphics[width=0.75\linewidth]{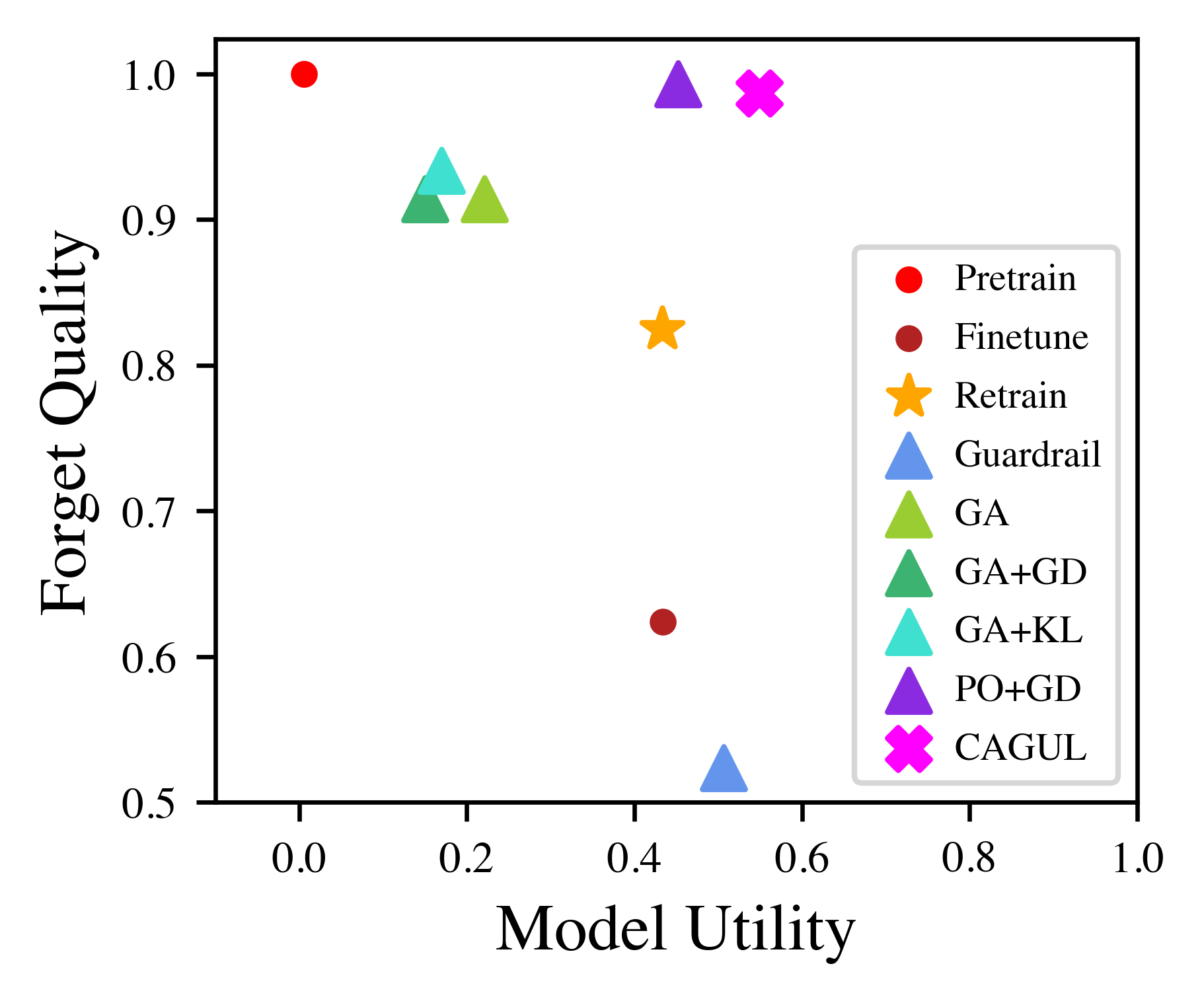}
    \caption{Unlearning performance comparison for \textit{LLaMA-3.2-11B-Vision-Instruct}; forget quality is plotted as 1 - harmonic mean for visualization purposes, \textit{upper-right is optimal}}
    \label{fig:pareto_main}
\end{figure}
Based on the performance of unlearning methods, we make the following observations:
We include main results from our experiments in~\cref{fig:pareto_main} for \textit{LLaMA-3.2-11B-Vision-Instruct}. Here, \textit{Pretrain} refers to the VLM used as is, and \textit{Finetune} is the VLM trained on the full FIUBench dataset to ensure adequate memorization of the dataset, as indicated by increased Rouge and EM scores compared to \textit{Pretrain}. We also provide individual performance metrics in~\cref{tab:main} where we observe a noticeable decline in the model's generalization ability on MME \& POPE datasets as the VLM is finetuned only on FIUBench. 
Nonetheless, we use this finetuned model as the base for unlearning.

\paragraph{CAGUL achieves better forget quality than most baselines, including the optimal \textit{Retrain} method.} In~\cref{tab:main}, we observe large drops in all forget metrics on $\mathcal{D}_f$ for CAGUL compared to \textit{Finetune}, indicating effective forgetting. CAGUL also achieves low APE and MinK values, which shows that our method is fairly robust against these adversarial and inference attacks. Note that CAGUL only trains small external modules and does not update any of the pretrained parameters, unlike finetuning baseline methods. Generally, all finetuning baselines achieve better forget metrics than \textit{Retrain}, and CAGUL outperforms most of these methods.

\paragraph{CAGUL preserves model utility after unlearning, outperforming all baseline methods.} Our results in~\cref{tab:main} show retain metrics on the full retain set $\mathcal{D}_r$ and the non-private subset $\mathcal{D}_{np}$ corresponding to the individuals in $\mathcal{D}_f$. We observe that CAGUL successfully preserves the retain utility of the finetuned base model. For $\mathcal{D}_{np}$, CAGUL retains and improves knowledge about the non-private data while simultaneously forgetting the private information of the same individuals. The increase in $\mathcal{D}_{np}$ metrics can be attributed to the further memorization of retain knowledge during training of unlearning methods. 
Generally, CAGUL outperforms baselines, including \textit{Retrain}, which suggests that the information encoded in visual tokens markedly helps to preserve retain performance.

\paragraph{CAGUL preserves the base model's generalization ability for downstream tasks.} Our results show that CAGUL retains model performance on both MME and POPE datasets, whereas baselines noticeably degrade it. We perform this comparison based on \textit{Finetune} performance, which serves as the minimum threshold for post-unlearning model behavior on non-forget data. Provided that the classifier can accurately distinguish images contained in the forget set, CAGUL ensures that pretrained knowledge is not manipulated in inference. 

\begin{table}[t]
    \centering
    \footnotesize
    \caption{Computational efficiency of CAGUL compared with baselines for \textit{LLaMA-3.2-11B-Vision-Instruct}}
    \begin{tabular}{l|c|c}
    \toprule
    \textbf{Method} & \textbf{Trainable Params} & \textbf{Exec. time (s)} \\
    \midrule
    Retrain      & $9.8$B & $4272$ \\
    GA           & $419$M &  $149$ \\
    GA+GD        & $419$M & $1855$ \\
    GA+KL        & $419$M & $2951$ \\
    PO+GD        & $419$M & $1911$ \\
    CAGUL        & $293$M &  $682$ \\
    \bottomrule
    \end{tabular}
    \label{tab:efficiency}
\end{table}
\paragraph{CAGUL has better computational efficiency than all finetuning-based unlearning methods.}~\cref{tab:efficiency} shows the number of training parameters and per epoch execution times for different unlearning methods with \textit{LLaMA-3.2-11B-Vision-Instruct}. \textit{Guardrail} method does not require any training as it performs inference-time prompt manipulation, but is ineffective at unlearning. Finetuning-based baselines require fewer parameters than \textit{Retrain}, as we implement LoRA finetuning for these methods as opposed to the full finetuning used for \textit{Retrain}. Nonetheless, CAGUL trains the fewest number of parameters, as the pretrained parameters are frozen, and completes training in less time. We note that GA requires the least training time as optimization is performed w.r.t. $\mathcal{D}_f$ only, but it causes the model to degrade quickly as indicated by its low model utility. During inference, we observe that CAGUL adds minimal latency to input processing, as the per-sample CPU time for CAGUL is approximately 0.29s, compared to baselines at 0.28s.

\begin{figure*}[t]
\centering
\begin{minipage}{.5\textwidth}
  \centering
  \includegraphics[width=0.95\linewidth]{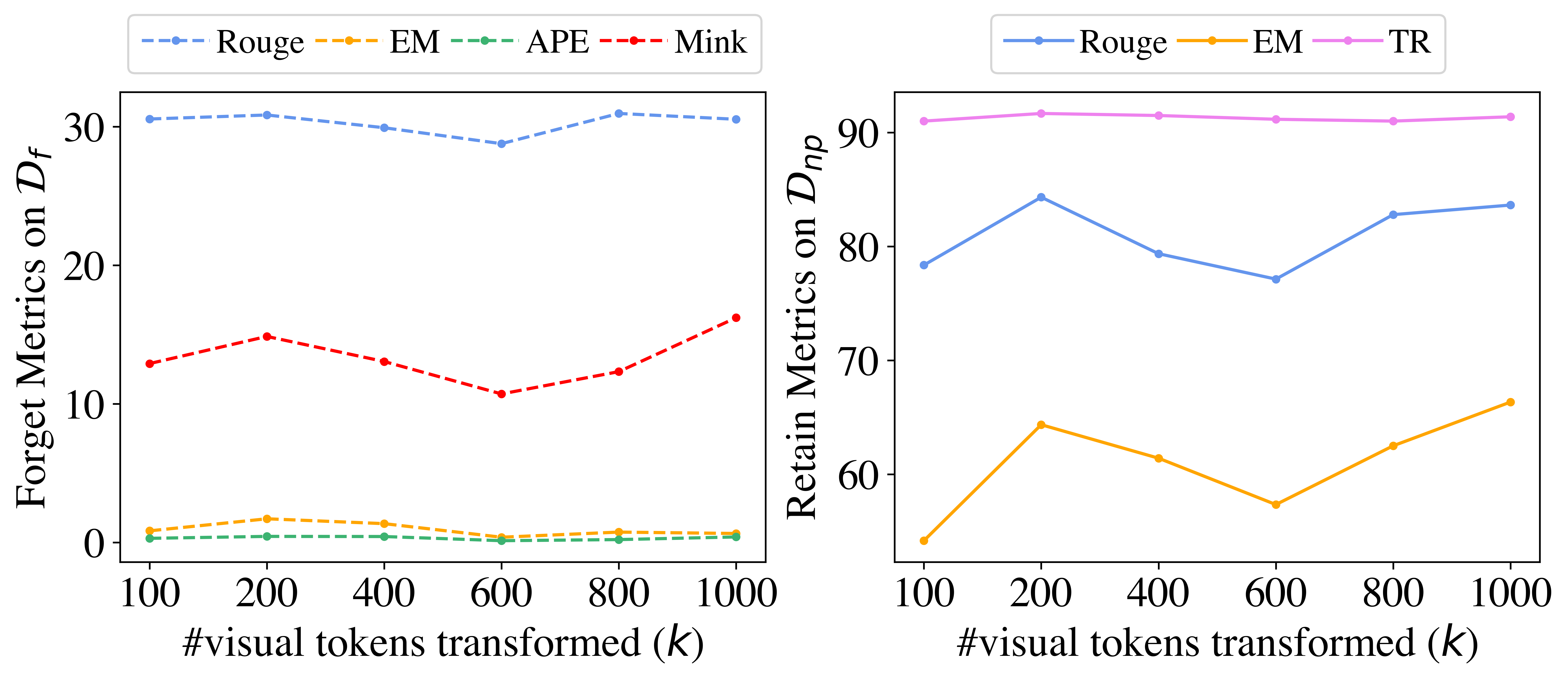}
\caption{Influence of $k$}
\label{fig:num_tokens}
\end{minipage}%
\begin{minipage}{.5\textwidth}
  \centering
  \includegraphics[width=0.95\linewidth]{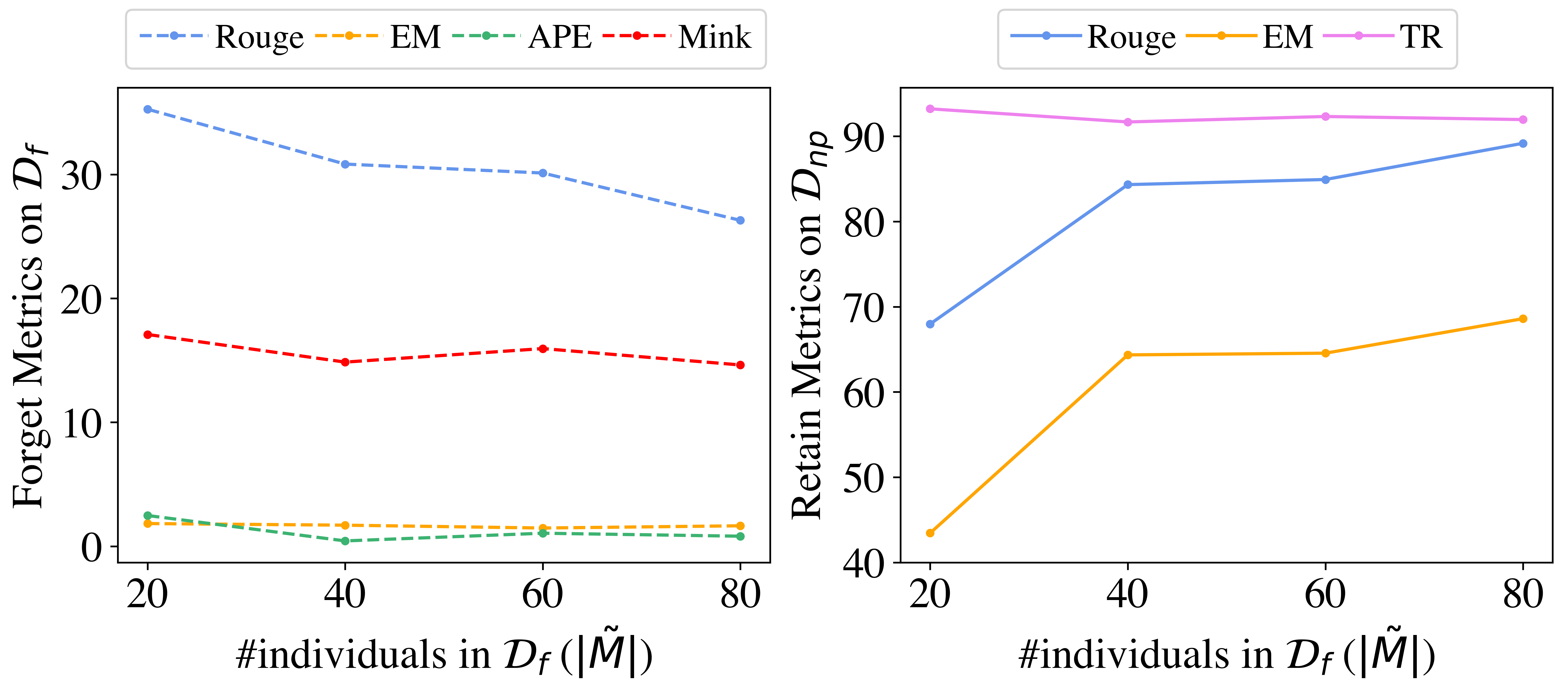}
\caption{Influence of $|\tilde{M}|$}
\label{fig:size}
\end{minipage}
\end{figure*}

\paragraph{CAGUL achieves better forget quality and model utility trade-offs with \textit{LLaMA-3.2-11B-Vision-Instruct} than \textit{Qwen-2.5-VL-7B-Instruct}.} We also run experiments with \textit{Qwen-2.5-VL-Instruct} whose details are included in~\cref{appendix:qwen}. We conjecture that this performance gap is due to the differing implementation of cross-modal attention in these models; specifically, the use of cross-attention in \textit{LLaMA} and joint self-attention in Qwen. 
Nonetheless, CAGUL with \textit{Qwen} achieves unlearning performance mostly comparable to retraining while also preserving downstream utility. 
Overall, our empirical results show that CAGUL achieves desirable trade-offs between forget quality and model utility 
comparable to SOTA baselines while saving on training costs.

\subsection{Additional Analysis}
\paragraph{Influence of Number of Visual Tokens Transformed.}
In CAGUL, we select the $k$ least attended visual tokens to encode unlearning-specific objectives via linear transformation using $\mathbf{F}_{\psi}$. Here, we study the influence of the value of $k$ on the overall performance of CAGUL by varying it as $\{100, 200, 400, 600, 800, 1000\}$. We report CAGUL performance on the forget set $\mathcal{D}_f$ and non-private retain set $\mathcal{D}_{np}$ in~\cref{fig:num_tokens}. We omit the results on $\mathcal{D}_r$, MME, and POPE datasets as all values of $k$ achieve the same metrics on these sets. 
Our results show peak performance for both forget and retain metrics when $k$ is set to $200$. We observe a minor dip at $k$ = $600$, but CAGUL generally achieves similar performance across different $k$ values.

\paragraph{Influence of Forget Set Size.}
We also study the effect of the size of forget set $\mathcal{D}_f$, which relates to the number of individuals participating in $\tilde{M}$. We vary $|\tilde{M}|$ across \{$20, 40, 60, 80$\} with corresponding dataset statistics shown in~\cref{tab:dataset_stat} and report forget and retain metrics in~\cref{fig:size}.
We observe a decreasing trend for the forget metrics on $\mathcal{D}_f$ and an increasing trend for the retain performance on $\mathcal{D}_{np}$ as $|\tilde{M}|$ becomes larger. This result suggests that unlearning with a larger forget set size facilitates more effective forgetting of private data and retaining of non-private data, most likely due to the increased size of the training subset used to learn visual token encoder weights.

\paragraph{Ablation Study.}
To understand the role of different components in CAGUL, we evaluate its variations as follows. We train CAGUL w/o $\mathbf{C}_\phi$ and $\mathbf{F}_\psi$ by removing the discriminator and visual token encoder, respectively. We evaluate CAGUL w/zero-out, where the embeddings for visual tokens selected via cross-attention are zeroed out instead of being transformed by the encoder $\mathbf{F}_\psi$. To understand the influence of attention-based token selection, we implement a variant of CAGUL w/random select, where the $k$ visual tokens are randomly chosen. Additionally, we run experiments where we substitute the loss in CAGUL with other commonly used unlearning losses: GA+GD and GA+KL. We present aggregate results from these experiments in~\cref{fig:pareto_ablation} and discuss individual metrics in~\cref{appendix:ablation}. Results show that the CAGUL achieves the best performance compared to its variants. CAGUL w/o$\mathbf{C}_\phi$ and w/random select have the least loss in performance, whereas CAGUL w/o$\mathbf{F}_\psi$ and w/zero-out have a large decline in forget performance, signifying that encoding unlearning signals in visual tokens through an external encoder is a promising unlearning strategy. Furthermore, the use of PO+GD loss is optimal for CAGUL as the other two unlearning losses cause a decline in either model utility or forget quality. 

\begin{figure}[t]
  \centering
    \includegraphics[width=0.7\linewidth]{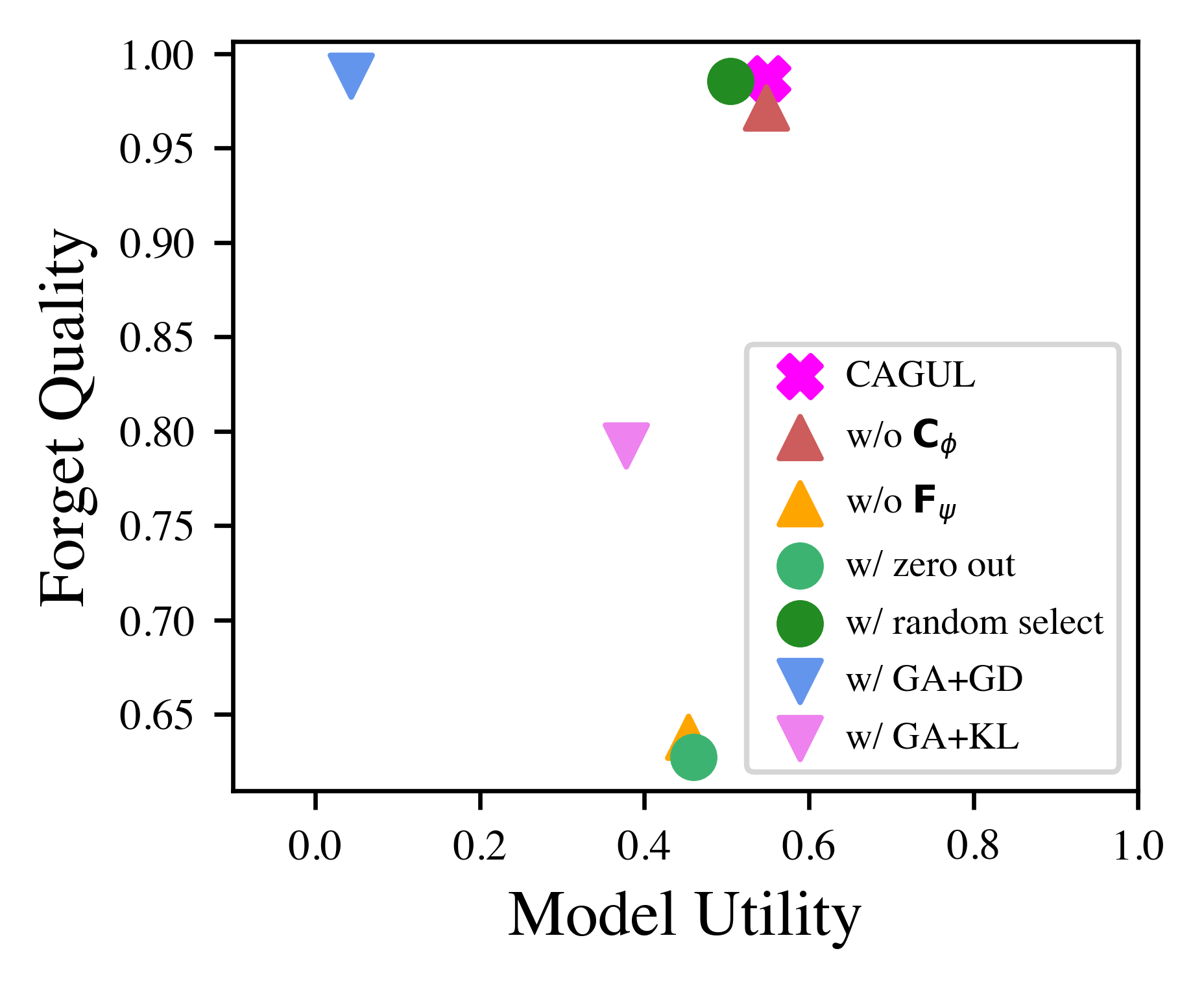}
    \caption{Overall results from ablation study}
    \label{fig:pareto_ablation}
\end{figure}

\section{Conclusion}
In this work, we proposed a novel setting for VLM unlearning where users can request the removal of a subset of their sensitive queries. To achieve unlearning, we proposed CAGUL framework, consisting of a classifier to identify images appearing in the forget set, a cross-modal attention-based visual token selection mechanism, and an encoder to embed unlearning information in visual tokens having low correlation with the query prompt. We conducted experiments on the FIUBench dataset using two open-source vision-language models and demonstrated the effectiveness of our method.

\section*{Acknowledgements}
This work is supported in part by NSF 1920920, 1946391, 2119691, 2147375, NIH P20GM139768, and the AHPCC.

\bibliographystyle{IEEEtran}
\bibliography{main}

\appendix

\subsection{Implementation Details}\label{appendix:implementation}

\paragraph{Cross-Modal Attention in Vision-Language Models.} We run our experiments using two state-of-the-art VLMs: \textit{LLaMA-3.2-11B-Vision-Instruct} and \textit{Qwen-2.5-VL-7B-Instruct}. The cross-modal alignment mechanism in VLMs is generally implemented as either \textit{cross-attention} or \textit{joint self-attention}. For cross-attention based models like \textit{LLaMA-3.2-11B-Vision-Instruct}, the query matrix $\bq$ and key matrix $\bk$ represent the two modalities separately, so $\ba$ directly captures the attention scores between the two modalities. In this scenario, each text token embedding is updated based on cross-attention scores computed \textit{before} input to the language model. However, for joint self-attention-based models such as \textit{Qwen-2.5-VL-7B-Instruct}, $\bz = (\bz_v, \bz_q)$ is fed in jointly as one concatenated input sequence to the language model, where $\bq = \bz\bw_q$ and $\bk = \bz\bw_k$ are shared among all tokens. The self-attention scores can be represented by the following block matrix:
\begin{equation}
\ba_{\text{self-attn}} = 
    \begin{bmatrix}
        \ba_{vv} & \ba_{vq} \\
        \ba_{qv} & \ba_{qq}
    \end{bmatrix},
\end{equation}
assuming the input is composed of an image followed by the query text for simplicity. For our formulation, we can consider $\ba$ for joint self-attention models to be the top right block $\ba_{vq}$ derived from $\bq = \bz_q\bw_q$ and $\bk = \bz_v\bw_k$, which represents the cross-modal attention scores between visual and text tokens. The cross-modal attention scores can be similarly extracted from the overall attention matrix based on the location of the image and query text in the input sequence. In CAGUL implementation, we obtain the cross-modal attention scores from the cross-attention scores of the final cross-attention layer in \textit{LLaMA-3.2-11B-Vision-Instruct} and joint self-attention scores of the last hidden layer in \textit{Qwen-2.5-VL-7B-Instruct}. The attention scores are averaged across attention heads and the query tokens to obtain aggregates for visual token selection in our method. As the visual token selection mechanism only utilizes query tokens, we can obtain least attended image tokens during inference for unseen samples as well.

\paragraph{Encoder Architectures.} We implement the discriminator $\mathbf{C}_{\phi}$ using a convolutional neural network and the visual token encoder $\mathbf{F}_{\psi}$ as a single-layer MLP. 

\begin{table*}[htb]
    \centering
    \caption{Training hyperparameters for finetuning and unlearning methods; parameters not applicable to method are shown as -}
    \resizebox{0.9\textwidth}{!}{
    \begin{tabular}{c|c|cccc|c}
    \toprule
        \textbf{Hyperparameters} & \textbf{Finetune/Retrain} & \textbf{GA} & \textbf{GA+GD} & \textbf{GA+KL} & \textbf{PO+GD} & \textbf{CAGUL} \\
    \midrule
        Learning rate & $2\times 10^{-5}$ & $1\times 10^{-6}$ & $2\times 10^{-5}$ & $1\times 10^{-4}$ & $3 \times 10^{-4}$ & $2\times 10^{-5}$ \\
        Batch size & $4$ & \multicolumn{4}{c|}{4} & $4$ \\
        Epochs & $10$ & \multicolumn{4}{c|}{$10$} & $2+10$ \\
        Dropout & - & \multicolumn{4}{c|}{$0.05$} & - \\
        LoRA Rank $r$ & - & \multicolumn{4}{c|}{$128$} & - \\
        LoRA Alpha $\alpha$ & - & \multicolumn{4}{c|}{$256$} & - \\
    \bottomrule
    \end{tabular}}
    \label{tab:hyperparameters}
\end{table*}

\paragraph{Setup} 
Following the norm in unlearning literature for pretrained models, we first finetune VLMs on the entire FIUBench dataset $\mathcal{D}$ to simulate data memorization. We conduct experiments with two VLMs: \textit{LLaMA-3.2-11B-Vision-Instruct} and \textit{Qwen-2.5-VL-7B-Instruct}. We implement full finetuning of the language and cross-modal components for both VLMs, i.e., the language model and the projector module for \textit{LLaMA}, and the language model and visual merger module for \textit{Qwen}. We refer to this finetuned VLM as the base model for unlearning. The objective here is to obtain an unlearned model which demonstrates forgetting of the personal information in $\mathcal{D}_f$ while maintaining the base model's performance for samples in $\mathcal{D}_r$, including $\mathcal{D}_{np}$.
We assume that $|\tilde{M}|$ = 40 individuals request removal of their private information, and report the main experiment results with \textit{LLaMA-3.2-11B-Vision-Instruct} as the pretrained VLM. We implement LoRA finetuning for baseline methods except retraining. For CAGUL, we train only the discriminator and visual token encoder while keeping the base model frozen as discussed in~\cref{sec:main}. More specifically, we first train the discriminator for 2 epochs, then jointly train the two modules and report results with perturbations performed on $k$ = 200 least attended visual tokens out of a total 6404 for \textit{LLaMA-3.2-11B-Vision-Instruct} and $k$ = 20 least attended visual tokens out of 1369 tokens for \textit{Qwen-2.5-VL-7B-Instruct}.   

Specific training hyperparameters for CAGUL and baselines are included in~\cref{tab:hyperparameters}. We use 4 NVIDIA A100 (40GB) GPUs for \textit{LLaMA-3.2-11B-Vision-Instruct} and 4 NVIDIA H200 (150GB) GPUs for \textit{Qwen-2.5-VL-7B-Instruct}  and report performance over a single run due to the resource-intensive nature of the experiments. Models are obtained directly from HuggingFace.

\subsection{Evaluation Metrics}\label{appendix:metrics}
\subsubsection{Forget Metrics}
\paragraph{Rouge-L.} We compute ROUGE-L scores to measure the similarity between generated text and ground truth answers for samples in $\mathcal{D}_f$ with lower Rouge being indicative of forgetting.

\paragraph{Exact Match.} Following~\cite{ma2025benchmarking}, we compute Exact Match (EM) scores to quantify the correctness of generated outputs compared to the ground truth labels. We obtain EM as an average over the ratio of ground-truth keywords appearing in the generated text for each query. For $\mathcal{D}_f$, a lower EM is desirable.

\paragraph{Adversarial Privacy Extraction.} Safety regulations in pre-trained models can often be bypassed by rephrasing query texts. The Adversarial Privacy Extraction (APE) was formulated to verify whether forgotten knowledge can be extracted from an unlearned model in an adversarial manner by prompting it with paraphrased queries~\cite{ma2025benchmarking}. We compute this metric as the average EM score when the unlearned model is queried with multiple paraphrases of each forget sample.

\paragraph{MinK.} Membership Inference Attacks (MIA) are often used to evaluate the forgetting effectiveness of unlearning methods. We leverage Min-K\% Prob~\cite{shi2023detecting} to quantify the presence of knowledge from $\mathcal{D}_f$ in the unlearned model. To compute this metric, we first obtain the probability for each generated token and calculate the average log likelihood over the first K\% tokens with minimum probabilities. A lower average log likelihood indicates that $\mathcal{D}_f$ was not included in the training data, demonstrating effective forgetting.

\subsubsection{Retain Metrics}

\paragraph{Rouge-L.} We compute ROUGE-L scores over the entire retain set $\mathcal{D}_r$. The retain objective is to preserve the base model's performance of this metric. For a thorough evaluation of forget and retain performance trade-off for the $\tilde{M}$ individuals requesting removal, we also provide the ROUGE-L scores for the retain subset $\mathcal{D}_{np}$ composed of non-private queries of individuals included in $\mathcal{D}_f$.

\paragraph{Exact Match.} For $\mathcal{D}_{np}$, we compute an EM score to evaluate whether the generated outputs from the unlearned model contain keywords specific to the ground truth to quantify CAGUL's ability to preserve utility.

\paragraph{Truth Ratio.}
We follow~\cite{ma2025benchmarking} and compute Truth Ratio (TR), which measures the model's tendency to generate factually incorrect answers versus correct ones. The likelihood of factual generation is computed as the probability of a paraphrased version of the ground truth answer, and the likelihood of an incorrect answer is calculated as an average over the probabilities of multiple perturbed answers formatted like the paraphrased answer. TR is reported as the ratio of incorrect to factual likelihoods. 

\subsubsection{General Downstream Accuracy}
We report accuracy scores on two general VQA datasets: MME~\cite{fu2023mme} and POPE~\cite{li2023pope} to demonstrate the influence of unlearning methods on the model's generalization ability. The MME benchmark is composed of queries related to various tasks such as existence, count, position, color, posters, celebrities, scenes, landmarks, and artworks. The POPE benchmark quantifies object hallucinations in VLMs, and we evaluate on the adversarial, popular, and random tasks.  

\subsubsection{Efficiency}
We additionally report per-epoch training time required by each unlearning method as well as the number of parameters that are updated during training to compare the computational efficiency of the implemented methods. We also consider the CPU time required during inference for each unlearning method.

\subsection{Additional Results}\label{appendix:addn_results}
\begin{table*}[t]
    \centering
    \caption{Fine-grained results from ablation study on \textit{LLaMA-3.2-11B-Vision-Instruct}}
    \resizebox{0.9\textwidth}{!}{
    \begin{tabular}{l|cccc|c|ccc|c|c}
    \toprule
    \multirow{4}{*}{\textbf{Method}} & \multicolumn{4}{c|}{\textbf{Forget}} & \multicolumn{4}{c|}{\textbf{Retain}} & \multicolumn{2}{c}{\textbf{General}} \\
    \cmidrule{2-11}
        & \multicolumn{4}{c|}{$\mathcal{D}_f$} & \multicolumn{1}{c|}{$\mathcal{D}_r$} & \multicolumn{3}{c|}{$\mathcal{D}_{np}$} & \multicolumn{1}{c|}{MME} & \multicolumn{1}{c}{POPE} \\
        \cmidrule{2-11}
        & Rouge$(\downarrow)$ & EM$(\downarrow)$ & APE$(\downarrow)$ & MinK$(\downarrow)$ & Rouge$(\uparrow)$ & Rouge$(\uparrow)$ & EM$(\uparrow)$ & TR$(\uparrow)$ & Acc.$(\uparrow)$ & Acc.$(\uparrow)$ \\
    \midrule
    CAGUL                & $30.84$ &  $1.70$ &  $0.43$ & $14.86$ & $85.15$ & $84.32$ & $64.35$ & $91.67$ & $30.58$ & $37.70$ \\
    \midrule
    w/o $\mathbf{C}_{\phi}$ & $30.94$ &  $1.84$ &  $1.28$ & $22.93$ & $96.07$ & $84.66$ & $65.94$ & $92.21$ & $28.94$ & $37.41$ \\
    w/o $\mathbf{F}_{\psi}$ & $56.56$ & $37.75$ & $23.67$ & $41.54$ & $84.50$ & $65.46$ & $28.51$ & $90.39$ & $30.58$ & $37.70$ \\
    w/ zero out                    & $57.12$ & $38.94$ & $24.47$ & $42.88$ & $84.53$ & $66.18$ & $29.89$ & $90.39$ & $30.58$ & $37.70$ \\
    \midrule
    w/ random select                & $31.23$ &  $1.70$ &  $0.48$ & $12.34$ & $84.71$ & $71.58$ & $43.45$ & $90.91$ & $30.58$ & $37.70$ \\
    \midrule
    w/ GA+GD                & $21.35$ &  $1.01$ &  $0.61$ &  $1.32$ & $82.99$ & $21.57$ &  $0.79$ & $77.76$ & $30.58$ & $37.70$ \\
    w/ GA+KL                & $53.86$ & $20.01$ & $12.61$ & $22.34$ & $84.47$ & $64.44$ & $16.60$ & $87.85$ & $30.58$ & $37.70$ \\
    \bottomrule
    \end{tabular}}
    \label{tab:ablation}
\end{table*}
\subsubsection{Ablation Study}\label{appendix:ablation}
We present the detailed results, including specific forget, retain, and downstream performance metrics from our ablation study in~\cref{tab:ablation} and corresponding discussion in the following.

\paragraph{Ablating Discriminator and Encoder.} First, we investigate the influence of the discriminator $\mathbf{C}_\phi$ and visual token encoder $\mathbf{F}_\psi$. We implement two versions of CAGUL, w/o $\mathbf{C}_\phi$ and w/o $\mathbf{F}_\psi$, where we remove the discriminator and the visual token encoder, respectively. For CAGUL w/o $\mathbf{C}_\phi$, we train only $\mathbf{F}_\psi$ and transform $k$ selected visual tokens for all $\mathcal{D}_f$ and $\mathcal{D}_r$, and for CAGUL w/o $\mathbf{F}_\psi$, we only train $\mathbf{C}_\phi$ and add uniform random noise to the $k$ visual tokens for the samples identified by $\mathbf{C}_\phi$. We also train a variant w/ zero out where we replace the embeddings for $k$ selected visual tokens with zeroes. Experiment results indicate that CAGUL can achieve good trade-offs between forget and retain metrics without the discriminator, but its generalization ability is negatively affected. We note that the rouge score for $\mathcal{D}_r$ shows noticeable improvement as the encoder is trained on the full retain set, enabling the model to memorize additional knowledge. Conversely, we observe a large decline in forget and retain performance for CAGUL w/o $\mathbf{F}_\psi$ and w/ zero out, which shows the importance of the encoder $\mathbf{F}_\psi$ in our framework.

\paragraph{Cross-Modal Attention Selection Strategy.} Another important component in CAGUL is the cross-modal attention-guided visual token selection for perturbation. Based on the reasoning that highly attended visual tokens provide most of the signals to the language module, we choose to perturb the $k$ least attended tokens with $\mathbf{F}_\psi$. We implement a variant of CAGUL where the $k$ visual tokens to be perturbed are randomly chosen for each sample sent to the encoder and report the results under w/ random select. The random selection of visual tokens achieves mostly comparable performance for forget metrics but underperforms for retain metrics, most notably in terms of Rouge and EM. It is likely that some highly attended visual tokens may get selected randomly, causing important visual signals to get distorted in the encoder. These results validate the selection of the least attended tokens for perturbation.

\begin{table*}[htb]
    \centering
    \caption{Unlearning metrics of CAGUL compared with baselines on \textit{Qwen-2.5-VL-7B-Instruct}}
    \resizebox{0.9\textwidth}{!}{
    \begin{tabular}{l|cccc|c|ccc|c|c}
    \toprule
    \multirow{4}{*}{\textbf{Method}} & \multicolumn{4}{c|}{\textbf{Forget}} & \multicolumn{4}{c|}{\textbf{Retain}} & \multicolumn{2}{c}{\textbf{General}} \\
    \cmidrule{2-11}
        & \multicolumn{4}{c|}{$\mathcal{D}_f$} & \multicolumn{1}{c|}{$\mathcal{D}_r$} & \multicolumn{3}{c|}{$\mathcal{D}_{np}$} & \multicolumn{1}{c|}{MME} & \multicolumn{1}{c}{POPE} \\
        \cmidrule{2-11}
        & Rouge$(\downarrow)$ & EM$(\downarrow)$ & APE$(\downarrow)$ & MinK$(\downarrow)$ & Rouge$(\uparrow)$ & Rouge$(\uparrow)$ & EM$(\uparrow)$ & TR$(\uparrow)$ & Acc.$(\uparrow)$ & Acc.$(\uparrow)$ \\
    \midrule
    Pretrain             & $30.61$ & $ 0.98$ & $ 0.37$ & $0.00$ & $36.27$ & $34.01$ & $ 0.10$ & $11.28$ & $86.68$ & $86.94$ \\
    Finetune             & $74.07$ & $56.92$ & $45.62$ & $2.83$ & $72.49$ & $68.18$ & $53.51$ & $89.85$ & $67.58$ & $79.24$ \\
    \midrule
    Retrain              & $44.04$ & $14.52$ & $10.70$ & $1.44$ & $72.68$ & $67.80$ & $54.23$ & $91.08$ & $71.31$ & $76.41$ \\
    Guardrail            & $52.34$ & $44.67$ & $35.61$ & $3.52$ & $70.53$ & $65.31$ & $45.17$ & $89.75$ & $66.94$ & $66.32$ \\
    GA                   & $ 0.02$ & $ 0.18$ & $ 0.18$ & $0.00$ & $ 0.04$ & $ 0.00$ & $ 0.00$ & $82.56$ & $28.99$ & $45.74$ \\
    GA+GD                & $45.39$ & $20.37$ & $13.92$ & $0.58$ & $76.88$ & $71.05$ & $78.47$ & $93.96$ & $75.72$ & $87.02$ \\
    GA+KL                & $ 4.66$ & $ 0.37$ & $ 0.18$ & $6.35$ & $ 8.66$ & $12.65$ & $ 0.79$ & $88.53$ & $68.68$ & $78.93$ \\
    PO+GD                & $42.41$ & $51.32$ & $30.63$ & $4.82$ & $81.08$ & $78.41$ & $96.30$ & $94.42$ & $72.42$ & $86.89$ \\
    CAGUL                & $40.82$ & $37.13$ & $23.78$ & $3.14$ & $71.88$ & $50.23$ & $40.01$ & $91.56$ & $68.64$ & $79.62$ \\
    \bottomrule
    \end{tabular}}
    \label{tab:qwen}
\end{table*} 

\paragraph{Choice of Forget and Retain Loss.} As discussed in Sec. 5 of the main paper, we train the encoder $\mathbf{F}_\psi$ using a combination of PO and GD as the forget and retain losses, respectively. We additionally run experiments where we substitute the loss in CAGUL with other commonly used unlearning losses: GA + GD and GA + KL. Similar to the baseline implementations of these losses, we utilize early stopping due to the GA loss and include results in~\cref{tab:ablation}. We observe that CAGUL w/ GA + GD achieves better forget metrics, but its retain performance quickly degrades despite early stopping. CAGUL w/ GA + KL relatively preserves retain performance but does not achieve satisfactory forget metrics, thus justifying the choice of PO + GD as the unlearning loss in CAGUL. Overall, the ablation results demonstrate the importance of the different components in the CAGUL formulation.

\begin{figure}[t]
    \centering
    \includegraphics[width=0.7\linewidth]{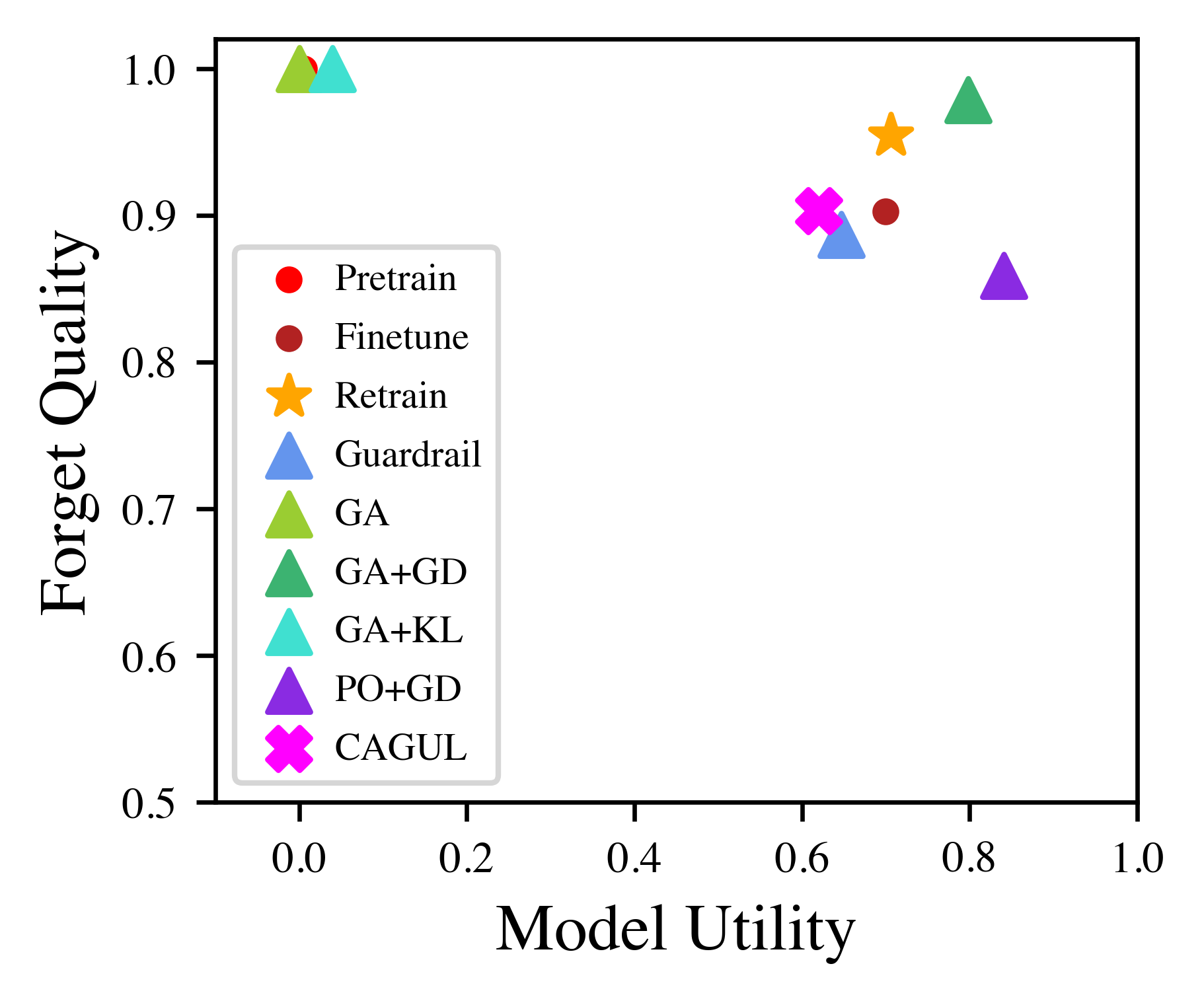}
    \caption{CAGUL aggregate performance comparison against baselines on \textit{Qwen-2.5-VL-7B-Instruct}}
    \label{fig:qwen}
\end{figure}

\subsubsection{Results on Qwen-2.5-VL-7B-Instruct}\label{appendix:qwen}
We show the results from our experiments on \textit{Qwen-2.5-VL-7B-Instruct} in~\cref{fig:qwen} and report individual metrics in~\cref{tab:qwen}. 
We first finetune \textit{Qwen-2.5-VL-7B-Instruct} on the FIUBench dataset to ensure sufficient memorization of individual profiles from the dataset, which is demonstrated by the increased performance metrics across $\mathcal{D}_f, \mathcal{D}_r$, and $\mathcal{D}_{np}$ for Finetune compared to Pretrain. We observe that, compared to \textit{LLaMA-3.2-11B-Vision-Instruct}, \textit{Qwen-2.5-VL-7B-Instruct} preserves downstream model performance on MME and POPE to a larger degree after finetuning. 

With CAGUL, we achieve satisfactory unlearning results as most metrics measured on $\mathcal{D}_f$ incur noticeable drops compared to Finetune. CAGUL also preserves the overall retain performance on $\mathcal{D}_r$ and the model's general utility on MME and POPE datasets. However, we notice a decline in the Rouge and EM scores for $\mathcal{D}_{np}$ despite an increase in the TR metric. We conjecture that this loss in performance arises as a result of the VLM's architecture. \textit{LLaMA-3.2-11B-Vision-Instruct} implements cross-attention layers that feed image representations into the language model discontinuously (i.e., only at a few select layers), whereas \textit{Qwen-2.5-VL-7B-Instruct} directly feeds image representations to the language model at every layer and computes self-attention for concatenated visual and text tokens instead of explicit cross-attention scores. As a result, encoding additional information in the visual tokens has an increased impact on generated outputs with \textit{Qwen}.

Among the baselines, although GA and GA+KL achieve superior forget performance, these methods seriously degrade model utility despite the use of early stopping. In contrast, CAGUL mostly retains model performance due to the use of the discriminator module $\mathbf{C}_{\phi}$, which only allows samples associated with forget images to be transformed by the encoder $\mathbf{F}_{\psi}$. Unlike \textit{LLaMA} results, GA+GD achieves the best trade-off for \textit{Qwen}, which is comparable to the Retrain method. Nonetheless, CAGUL achieves better forget metrics than its baseline finetuning counterpart PO+GD, while preserving the overall retain utility and general downstream accuracy. Note that all finetuning baseline methods implement model finetuning and require more computational resources than CAGUL, which achieves satisfactory unlearning performance while keeping the VLM frozen.

\end{document}